\documentclass{article}
 \usepackage[left=1.5in,top=1.25in,right=1.5in,bottom=1.25in,head=1.25in]{geometry}
\usepackage{amsmath,amsthm,amssymb,graphicx,url,natbib}

\usepackage{times, array, pifont,color}

\newcommand{\Hc}{\mathcal{H}}

\newcommand{\tphi}{\widetilde{\phi}}
\newcommand{\tpsi}{\widetilde{\psi}}
\newcommand{\HSIC}{\textrm{HSIC}}
\newcommand{\tr}{\textrm{tr}}

\newcommand{\SigXY}{\Sigma_{XY}}
\newcommand{\SXY}{S_{XY}}

\newcommand{\tL}{\widetilde{L}}
\newcommand{\tK}{\widetilde{K}}
\newcommand{\tl}{\widetilde{l}}
\newcommand{\tk}{\widetilde{k}}

\def\E{\mathbb{E}}
\newtheorem{proposition}{Proposition}
\DeclareMathOperator*{\argmin}{arg\,min}

\newcommand{\cmark}{\ding{51}}%
\newcommand{\xmark}{\ding{55}}%

\title{Nonparametric Independence Testing for \\ Small Sample Sizes}

\author{
Aaditya Ramdas\footnote{Both authors have equally contributed to this paper.} \\
Dept. of Statistics and Machine Learning Dept.\\
Carnegie Mellon University\\
aramdas@cs.cmu.edu\\
\and 
Leila Wehbe$^{*}$ \\
Machine Learning Dept.\\
Carnegie Mellon University\\
lwehbe@cs.cmu.edu\\
}

\begin{document}

\maketitle

\begin{abstract}
This paper deals with the problem of nonparametric independence testing, a fundamental decision-theoretic problem that asks if two arbitrary (possibly multivariate) random variables $X,Y$ are independent or not, a question that comes up in many fields like causality and neuroscience. While quantities like correlation of $X,Y$ only test for (univariate) linear independence, natural alternatives like mutual information of $X,Y$ are hard to estimate due to a serious curse of dimensionality. A recent approach, avoiding both issues, estimates norms of an \textit{operator} in Reproducing Kernel Hilbert Spaces (RKHSs). Our main contribution is strong empirical evidence that by employing \textit{shrunk} operators when the sample size is small, one can attain an improvement in power at low false positive rates. We analyze the effects of Stein shrinkage on a popular test statistic called HSIC (Hilbert-Schmidt Independence Criterion).
Our observations provide insights into two recently proposed shrinkage estimators, SCOSE and FCOSE - we prove that SCOSE is (essentially) the optimal linear shrinkage method for \textit{estimating} the true operator; however, the non-linearly shrunk FCOSE usually achieves greater improvements in \textit{test power}. This work is important for more powerful nonparametric detection of subtle nonlinear dependencies for small samples.
\end{abstract}


\section{Introduction}

The problem of \textit{nonparametric} independence testing deals with ascertaining if two random variables are independent or not, making no parametric assumptions about their underlying distributions. Formally, given $n$ samples $(x_i,y_i)$ for $i \in \{1,...,n\}$ where $x_i \in \mathbb{R}^p, y_i \in \mathbb{R}^q$, that are drawn from a joint distribution $P_{XY}$ supported on $\mathcal{X} \times \mathcal{Y} \subseteq \mathbb{R}^{p+q}$, we want to decide between the \textit{null} and \textit{alternate} hypotheses
$$
\Hc_0: P_{XY} = P_X \times P_Y ~\mbox{~vs.~}~ \Hc_1: P_{XY} \neq P_X \times P_Y
$$
where $P_X,P_Y$ are the marginals of $P_{XY}$ w.r.t. $X,Y$. A test is a function from the data to $\{0,1\}$. Tests aim to have high power (probability of detecting dependence, when it exists) at a prespecified allowable type-1 error rate $\alpha$ (probability of detecting dependence when there isn't any).

Independence testing is often a precursor to further analysis. Consider for instance conditional independence testing for inferring causality, say by the PC algorithm \cite{pc}, whose first step is (unconditional) independence testing. It is also useful for scientific discovery like in neuroscience, to see if a stimulus $X$ (say an image) is independent of the brain activity $Y$ (say fMRI) in a relevant part of the brain. Since \textit{detecting} nonlinear correlations is much easier than \textit{estimating} a nonparametric regression function (of $Y$ onto $X$), it can be done at smaller sample sizes, with further samples collected for estimation only if an effect is detected by the hypothesis test. For such situations, correlation only tests for univariate linear independence, while other statistics like mutual information that do characterize multivariate independence are hard to estimate from data, suffering from a serious curse of dimensionality. A recent popular approach for this problem (and a related two-sample testing problem) involve the use of quantities defined in reproducing kernel Hilbert spaces (RKHSs) - see \cite{mmd1,kfda,coco,hsic}. 

This paper will concern itself with increasing the statistical power at small samples of a popular kernel statistic called HSIC, by using \textit{shrunk} empirical estimators of the unknown population quantity (introduced below).

\subsection{Hilbert Schmidt Independence Criterion}

Due to limited space, familiarity with RKHS terminology is assumed - see \cite{learningkernels} for an introduction. Let $k : \mathcal{X} \times \mathcal{X} \rightarrow \mathbb{R}$ and $l : \mathcal{Y} \times \mathcal{Y} \rightarrow \mathbb{R}$ be two positive-definite reproducing kernels that correspond to RKHSs $\Hc_k$ and $\Hc_l$ respectively with inner-products $\langle \cdot,\cdot \rangle_k$ and $\langle \cdot , \cdot \rangle_l$. Let $k,l$  arise from (implicit) feature maps $\phi : \mathcal{X} \to \Hc_k$ and $\psi : \mathcal{Y} \to \Hc_l$. In other words, $\phi, \psi$ are not functions, but mappings to the Hilbert space. i.e. $\phi(x) \in \Hc_k,\psi(y) \in \Hc_l$ respectively. These functions, when evaluated at points in the original spaces, must satisfy $\phi(x)(x') = \langle \phi(x), \phi(x') \rangle_k= k(x,x')$ and $\psi(y)(y') =\langle \psi(y), \psi(y') \rangle_l= l(y,y')$.

The mean embedding of $P_X$ and $P_Y$ are defined as $\mu_X := \mathbb{E}_{x \sim P_X} \phi(x) \in \Hc_k$ and  $\mu_Y := \mathbb{E}_{y \sim P_Y} \psi(y) \in \Hc_l$ whose empirical estimates are $\widehat{\mu}_X := \frac1{n} \sum_{i=1}^n\phi(x_i)$ and $\widehat{\mu}_Y := \frac1{n} \sum_{i=1}^n \psi(y_i)$. Finally, the cross-covariance operator of $X,Y$ is defined as
$$
\Sigma_{XY} := \mathbb{E}_{(x,y) \sim P_{XY}} (\phi(x) - \mu_X) \otimes (\psi(y)-\mu_Y)
$$
where $\otimes$ is an outer-product. For unfamiliar readers, if we used the linear kernel $k(x,x')=x^T x'$ and $l(y,y') = y^T y'$, then the cross-covariance operator is just the cross-covariance matrix. The plug-in empirical estimator of $\Sigma_{XY}$ is

$$
S_{XY} := \frac1{n} \sum_{i=1}^n (\phi(x_i) - \widehat{\mu}_X) \otimes (\psi(y_i) - \widehat{\mu}_Y)
$$
For conciseness, define $\tphi(x_i) = \phi(x_i) - \widehat{\mu}_X$, $\tpsi(y_i) = \psi(y_i) - \widehat{\mu}_Y$, $\tk(x,x') = \langle \tphi(x), \tphi(x') \rangle_k$ and $\tl(y,y') = \langle \tpsi(y), \tpsi(y') \rangle_l$.
The test statistic Hilbert-Schmidt Independence Criterion (HSIC) defined in \cite{hsic} is the squared Hilbert-Schmidt norm of $\SXY$, and can be calculated using centered kernel matrices $\tK,\tL$, where $\tK_{ij}=\tk(x_i,x_j), \tL_{ij}=\tl(y_i,y_j)$, as
\begin{equation}
\HSIC := \|S_{XY}\|_{HS}^2 = \frac1{n^2} \tr(\tK \tL)
\end{equation}
For unfamiliar readers, if we used the linear kernel, this just corresponds to the Frobenius norm of the cross-covariance matrix. 
The most important property is: \textit{when the kernels $k,l$ are ``characteristic'', then the corresponding population statistic $\|\Sigma_{XY}\|_{HS}^2$ is zero iff  $X,Y$ are independent} \cite{hsic}. This gives rise to a natural test -  calculate $\|S_{XY}\|_{HS}^2$ and reject the null if it is large. 

Examples of characteristic kernels include Gaussian $k(x,x')=\exp\left(-\frac{\|x-x'\|_2^2}{\gamma^2}\right)$ and Laplace $k(x,x') = \exp\left(-\frac{\|x-x'\|_1}{\gamma} \right)$, for any bandwidth $\gamma$, while the aforementioned linear kernel is not characteristic --- the corresponding HSIC tests only linear relationships, and a zero cross-covariance matrix characterizes independence  only for multivariate Gaussian distributions. Working with the infinite dimensional operator with characteristic kernels, allows us to identify any general nonlinear dependence (in the limit) between any pair of distributions, not just Gaussians.

\subsection{Independence Testing using HSIC}

A permutation-based test is described in \cite{hsic}, and proceeds in the following manner.
From the given data, calculate the test statistic $T := \|S_{XY}\|_{HS}^2$. Keeping the order of $x_1,...,x_n$ fixed, randomly permute $y_1,...,y_n$ a large number of times, and recompute the \textit{permuted} HSIC each time. This destroyed any dependence between $x,y$ simulating a draw from the product of marginals, making the empirical distribution of the permuted HSICs behave like the null distribution of the test statistic (distribution of $\HSIC$ when $\Hc_0$ is true). For a pre-specified type-1 error $\alpha$, calculate threshold $t_\alpha$ in the right tail of the null distribution. Reject $\Hc_0$ if $T > t_\alpha$. 
This test was proved to be \textit{consistent} against any fixed alternative, meaning  for any fixed type-1 error $\alpha$, the power goes to 1 as $n \rightarrow \infty$. Empirically, the power can be calculated using simulations by repeating the above permutation test many times for a fixed $P_{XY}$ (for which dependence holds), and reporting the empirical probability of rejecting the null (detecting the dependence). Note that the power depends on $P_{XY}$ (unknown to the user of the test).


\subsection{Shrunk Estimators of $\SXY$}

Even though $S_{XY}$ is an unbiased estimator of $\Sigma_{XY}$, it typically has high variance at low sample sizes. The idea of Stein shrinkage \cite{Stein} is to trade-off bias and variance, first introduced in the context of Gaussian mean estimation. This strategy of   introducing some bias and decreasing the variance to get different estimators of $\Sigma_{XY}$  was followed by \cite{kernelstein} who define a linear shrinkage estimator of $S_{XY}$ called SCOSE (Simple Covariance Shrinkage Estimator) and a nonlinear shrinkage estimator called FCOSE (Flexible Covariance Shrinkage Estimator). When we refer to shrunk estimators, we implicitly mean SCOSE and FCOSE. We will describe these briefly in Section 2.

\subsection{Contributions}

 Our first contribution is the following :

1. We provide evidence that employing shrunk estimators of $\Sigma_{XY}$, instead of $S_{XY}$, to calculate the aforementioned test statistic, can increase the power of the associated independence test at low false positive rates, when the sample size is small (there is higher variance in estimating infinite-dimensional operators).

Our second contribution is to analyze the effect of shrinkage on the test statistic, to provide some practical insight.

2. The effect of shrinkage on the test-statistic is very similar to soft-thresholding (see Section 4), shrinking very small statistics to zero, and shrinking other values nearly (but not) linearly, and nearly (but not) monotonically. 

Our last contribution is an insight on the two estimators considered in this paper, SCOSE and FCOSE.

3. We prove that SCOSE is (essentially, up to lower order terms) the optimal/oracle linear shrinkage estimator with respect to quadratic risk (see Section 5). However, we observe that FCOSE typically achieves higher power than SCOSE. This indicates that it may be useful to search for the optimal estimator in a larger class than linearly shrunk estimators, and also that quadratic loss may not be the right loss function for the purposes of test power.

The rest of this paper is organized as follows. Section 2 introduces SCOSE, FCOSE and their corresponding shrunk test statistics. Section 3 presents illuminating experiments that bring out the statistically significant improvement in power over HSIC. Section 4 conducts a deeper investigation into the effect of shrinkage and proves the oracle optimality of SCOSE under quadratic risk.

\section{Shrunk Estimators and Test Statistics}

 Let $\mathcal{HS}(\Hc_k,\Hc_l)$ represent the set of Hilbert-Schmidt operators from $\Hc_k$ to $\Hc_l$. We first note that $S_{XY}$ can be written as the solution to the following optimization problem.
$$
S_{XY} := \min_{Z \in \mathcal{HS}(\Hc_k,\Hc_l)} \frac1{n} \sum_{i=1}^n \left \| \tphi(x_i)\otimes \tpsi(y_i) - Z \right\|_{HS}^2
$$
Using this idea \cite{kernelstein} suggest the following two shrunk/regularized estimators. 

\subsection*{From SCOSE to $\HSIC^S$}
This is derived in \cite{kernelstein} by solving 
$$
\min_{Z \in \mathcal{HS}(\Hc_k,\Hc_l)} \frac1{n} \sum_{i=1}^n \left \|\tphi(x_i)\otimes \tpsi(y_i) - Z \right\|_{HS}^2 + \lambda \|Z\|_{HS}^2
$$
and the optimal solution  (called SCOSE) is
$$
S_{XY}^{S} := \left(1 - \frac{\lambda}{1+\lambda}\right) S_{XY} 
$$
where $\lambda$ (and hence the shrinkage intensity) is estimated by leave-one-out cross-validation (LOOCV), in closed form as
\begin{align*}
\rho^S &:= \left(\frac{\lambda^{CV}}{1+\lambda^{CV}}\right) \\
&= \frac{\left[ \frac{1}{n} \sum_{i=1}^{n}  \tK_{ii}\tL_{ii}  - \frac{1}{n^2} \sum_{i,j=1}^{n} \tK_{ij}\tL_{ij} \right]}{(n-2)\frac{1}{n^2} \sum_{i,j=1}^{n} \tK_{ij}\tL_{ij} +  \frac{1}{n^2} \sum_{i=1}^{n}  \tK_{ii}\tL_{ii} }
\end{align*}
Observing the expression for $\lambda^{CV}$ in \cite{kernelstein}, the denominator can be negative (for example, with the Gaussian kernel for small bandwidths, resulting in a kernel matrix close to the identity). This can cause $\lambda^{CV}$ to be negative, and $\rho^S$ to be (unintentionally) outside the range $[0,1]$.  Though not discussed in \cite{kernelstein}, we shall follow the convention that when $\rho^S < 0$, we shall use $\rho^S = 0$ and if $\rho^S > 1$, we use $\rho_S = 1$. Indeed, one can show that $\left(1 - \frac{\lambda}{1+\lambda}\right)_+ S_{XY}$ dominates $\left(1 - \frac{\lambda}{1+\lambda}\right) S_{XY}$ where $(x)_+ = \max\{x,0\}$. In Section 4, we prove that $\SXY^S$ is (essentially) the optimal/oracle linear shrinkage estimator with respect to quadratic risk.

We can now calculate the corresponding shrunk statistic $\HSIC^S ~=~ \|\SXY^S \|_{HS}^2 ~=~$
\begin{equation}
\left(1 - \frac{\frac1{n}\sum_{i=1}^n\tK_{ii}\tL_{ii} - \HSIC}{(n-2)\HSIC +  \tfrac{\frac1{n}\sum_{i=1}^n\tK_{ii}\tL_{ii}}{n} } \right)_+^2  \HSIC ~\  ~\  ~ \label{eq:HSICS}
\end{equation}

While the above expression looks daunting, one thing to note is that the amount that $\HSIC$ is shrunk (i.e. the multiplicative factor) depends on the value of $\HSIC$. As we shall see in section 4, small $\HSIC$ values get shrunk to zero, but as can be seen above, the shrinkage of HSIC is non-monotonic. 

\subsection*{From FCOSE to $\HSIC^F$}
The Flexible Covariance Shrinkage Estimator is derived by relying on the Representer theorem, see \cite{learningkernels}, to instead minimize
$$
\frac1{n} \sum_{i=1}^n \left \|\tphi(x_i)\otimes \tpsi(y_i) - \sum_{i=1}^n \frac{\beta_i}{n} \tphi(x_i)\otimes \tpsi(y_i) \right\|_{HS}^2 + \lambda \|\beta\|_2^2
$$
over all $\beta \in \mathbb{R}^n$, and the optimal solution (called FCOSE) is
\begin{eqnarray*}
\SXY^{F} &:=& \sum_{i=1}^n \frac{\beta^\lambda_i}{n} \tphi(x_i)\otimes \tpsi(y_i) \\
~\mbox{where}~ \beta^\lambda &=& (\tK \circ \tL + \lambda I)^{-1} \tK \circ \tL \mathbf{1}
\end{eqnarray*}
where $\circ$ denotes elementwise (Hadamard) product, $\mathbf{1}$ is the vector $[1, 1, ..., 1]^T$, and as before the best $\lambda$ is determined by LOOCV. The procedure to evaluate the optimal $\lambda$ efficiently is described 
 by \cite{kernelstein} - a single eigenvalue decomposition of $\tK \circ \tL$ costing $O(n^3)$ can be done, following which evaluating LOOCV is only $O(n^2)$ per $\lambda$, see \cite{kernelstein}, section 3.1 for more details. As before, after picking the $\lambda$ by LOOCV, we can derive the corresponding shrunk test statistic as
\begin{align*}
\HSIC^{F} &= \|\SXY^S \|_{HS}^2 \\
&= \frac1{n^2} \tr(M(M+\lambda I)^{-1}M(M+\lambda I)^{-1}M)
\end{align*}

where $M = \tK \circ \tL$. Note here that the shrinkage is not linear, and the effect on $\HSIC$ cannot be seen immediately. Similar to SCOSE, we shall see in section 4, small $\HSIC$ values get shrunk to zero (LOOCV chooses a large $\lambda$).

\section{Linear Shrinkage and Quadratic Risk}

In this section, we prove that SCOSE is (essentially) optimal within a particular class of estimators. Such ``oracle'' arguments also exist elsewhere in the literature, like \cite{ledoitwolf}, so we provide only a brief proof outline.





\begin{proposition}
The oracle (with respect to quadratic risk) linear shrinkage estimator and intensity is defined as
\begin{eqnarray*}
S^*,\rho^* := \argmin_{Z \in \mathcal{HS}, Z = (1-\rho)\SXY, 0 \leq \rho \leq 1 } \|Z - \SigXY\|_{HS}^2
\end{eqnarray*}
and is given by $S^* := (1-\rho^*)\SXY$ where 
$$
\rho^* ~:=~ \frac{\E\|\SXY - \SigXY\|_{HS}^2}{\E\|\SXY\|^2}  
$$
\end{proposition}
\begin{proof}
Define $\alpha^2 = \|\SigXY\|_{HS}^2$, $\beta^2 = \E\|\SXY - \SigXY\|_{HS}^2$, $\delta^2 = \E\|\SXY\|^2$. Since $\E[\SXY] = \SigXY$, 
 it is easy to verify that
$\alpha^2  + \beta^2 = \delta^2$. 
Substituting and expanding the objective, we get:
\begin{eqnarray*}
\E\|Z - \SigXY\|_{HS}^2 &=& \E\|-\rho \SXY + (\SXY - \SigXY)\|_{HS}^2 \nonumber \\
 &=& \rho^2 \delta^2 + \beta^2  - 2\rho (\delta^2 - \alpha^2)\\
&=& \rho^2 \alpha^2 + (1-\rho)^2 \beta^2
\end{eqnarray*}
Differentiating and equating to zero, gives $\rho^* ~=~ \frac{\beta^2}{\delta^2}$.
\end{proof}


 This $\rho^*$ appears in terms of  quantities that depend on the unknown underlying distribution (hence  the term \textit{oracle} estimator). We use plugin estimates $b,d$ for $\beta,\delta$. Let $d^2~=~ \|\SXY\|_{HS}^2 ~=~ \frac1{n^2}\sum_{i,j=1}^n \tK_{ij}\tL_{ij} = HSIC$. Since $\beta^2$ is the variance of  $\SXY$, let $b^2$ be the sample variance of $\SXY$, i.e.  $b^2 = \frac{1}{n}\frac1{n} \sum_{k=1}^{n} || \tphi(x_i) \otimes \tpsi_{x_i} - \SXY||^2 = \frac1{n} \left[ \frac{1}{n} \sum_{i=1}^{n}  \tK_{ii}\tL_{ii}  - \frac{1}{n^2} \sum_{i,j=1}^{n} \tK_{ij}\tL_{ij} \right]$. Plugging these into $S^*$ and simplifying, we see that $\HSIC^* := \|S^*\|_{HS}^2 $ is
\begin{equation}
\HSIC^* = \left(1 - \frac{ \frac1{n}\sum_{i=1}^n\tK_{ii}\tL_{ii} - \HSIC}{n \HSIC} \right)^2 \HSIC
\label{eq:oraclehsic}
\end{equation}
Comparing Eq.\eqref{eq:oraclehsic} with Eq.\eqref{eq:HSICS} shows that SCOSE is essentially $S^*$, up to a factor in the denominator which is of the same order as the bias of the HSIC empirical estimator\footnote{$\HSIC$ and $\HSIC - 2\HSIC/n - C/n^2$ both converge to population HSIC at same rate determined by the dominant term ($\HSIC$).} (see Theorem 1 in \cite{hsic}).
In other words, SCOSE just corresponds to using a slightly different estimator for $\delta^2$ than the simple plugin $d^2$, which varies on the same order as the bias $\delta^2 - \E d^2$. Hence SCOSE, as estimated via regularization and LOOCV, is (essentially) the optimal linear shrinkage estimator under quadratic risk.

To the best of our knowledge, this is the first such characterization of \textit{optimality of an estimator achieved through leave-one-out cross-validation}. We are only able to prove this because one can explicitly calculate both the oracle linear shrinkage intensity $\rho^*$ as well as the optimal $\lambda^{CV}$ (as mentioned in Section 2). This raises a natural open question  --- can we find other situations where the LOOCV estimator is optimal with respect to some risk measure?  (perhaps when explicit calculations are not possible, like ridge regression).

\section{Experiments}

In this section, we run three kinds of experiments: a) to verify that SCOSE has better quadratic risk than FCOSE and original sample estimator, b) detailed synthetic experiments to verify that shrinkage does improve power, across interesting regimes of $\alpha = \{0.01,0.05,0.1\}$, and c) real data obtained from MNIST, to show that we shrinkage detect dependence at much lower samples than the original data size.

\subsection{Quadratic Risk}

\begin{figure} [h!]
\centering
\includegraphics[width=0.41\linewidth]{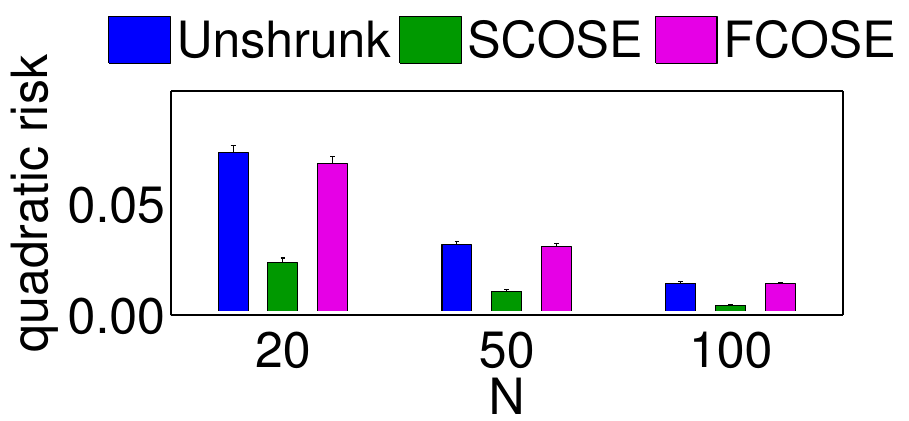}
\includegraphics[width=0.41\linewidth]{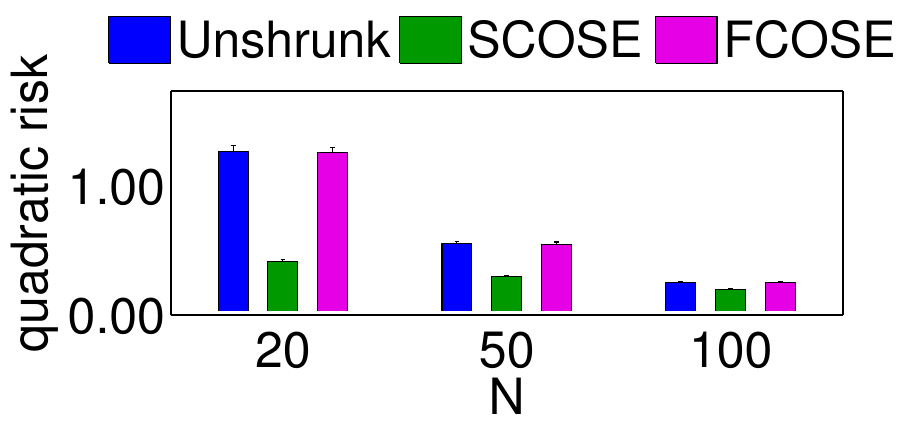}
\includegraphics[width=0.41\linewidth]{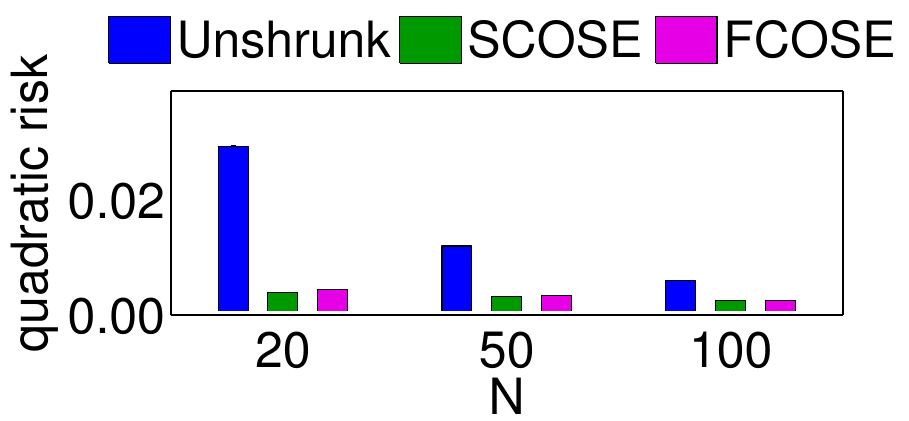}
\includegraphics[width=0.41\linewidth]{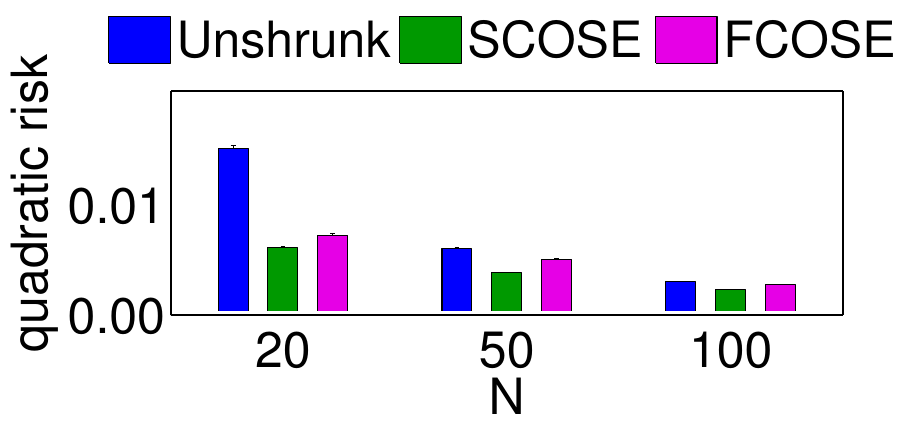}
\caption{All panels show quadratic risk $\E\|X-\SigXY\|_{HS}^2$ for $X\in\{\SXY,\SXY^{S},\SXY^{F}\}$. Dataset (A) was used in all four panels, but the kernels were varied - from top to bottom is the linear, quadratic, Gaussian and Laplace kernel.}
\label{fig:qrisk}
\end{figure}

Figure \ref{fig:qrisk} shows that SCOSE is indeed much better than both $\SXY$ and FCOSE with respect to quadratic risk. Here, we calculate $\E\|Z - \SigXY\|_{HS}^2$ for the distribution given in dataset (A) for $Z \in \{\SXY, \SXY^S, \SXY^F\}$. The expectation is calculated by repeating the experiment 1000 times. Each time $Z$ is calculated according to $N \in \{20,50,100\}$ samples and $\SigXY$ is approximated by the empirical cross-covariance matrix on 5,000 samples. The four panels use four different kernels which are linear, polynomial, Laplace  and Gaussian from top to bottom. The shrunk estimators are always better than the unshrunk, with a larger difference between SCOSE and FCOSE  for finite-dimensional feature spaces (top two). In infinite-dimensional feature spaces (bottom two), SCOSE and FCOSE are much better than the unshrunk estimator but very similar to each other. The differences between all estimators decreases with increasing $n$, since the sample cross-covariance operator itself becomes very accurate.

\subsection{Synthetic Data}

We perform synthetic experiments in a wide variety of settings to demonstrate that the shrunk test statistics achieve higher power than HSIC in a variety of settings. We follow the schema provided in the introduction for independence testing and calculating power. We only consider difficult distributions with nonlinear dependence between $X,Y$, on which linear methods like correlation are shown to fail to detect dependence (some of them were used in previous papers on independence testing like \cite{hsic2008} and \cite{timeseries}).
%
%
%
%
%

For all experiments, $\alpha \in \{0.01, 0.05, 0.1\}$ is chosen as the type-1 error (for choosing the threshold level of the null distribution's right tail). For every setting of parameters of each experiment, power is calculated as the percentage of rejection over 200 repetitions (independent trials), with 2000 permutations per repetition (permutation testing to find the null distribution threshold at level $\alpha$). We use the Gaussian kernel where the bandwidth is chosen by the common median heuristic \cite{learningkernels}.

Table \ref{tab:synth} is a representative sample from what we saw on other examples - either large, small or no improvement in power was seen but almost never a worsening of power. The improvements in power may not always be huge, but they are statistically significant - it is difficult to detect such non-linear dependencies at low sample sizes, so \textit{any} increase in power can be important in scientific applications.

\textbf{Remark.} A more appropriate way than using error bars to assess significance is by the Wilcoxon rank sum test, omitted for lack of space, though it yields more favorable results.

\subsection{Real Data}

We use two real datasets - the first is a good example where shrinkage helps a lot, but in the second it does not help (we show it on purpose). Like the synthetic datasets, for most real datasets it either helps or does not hurt (being very rarely worse; see remark in the discussion section).

\begin{figure} [h!]
\centering
\includegraphics[width=0.41\linewidth]{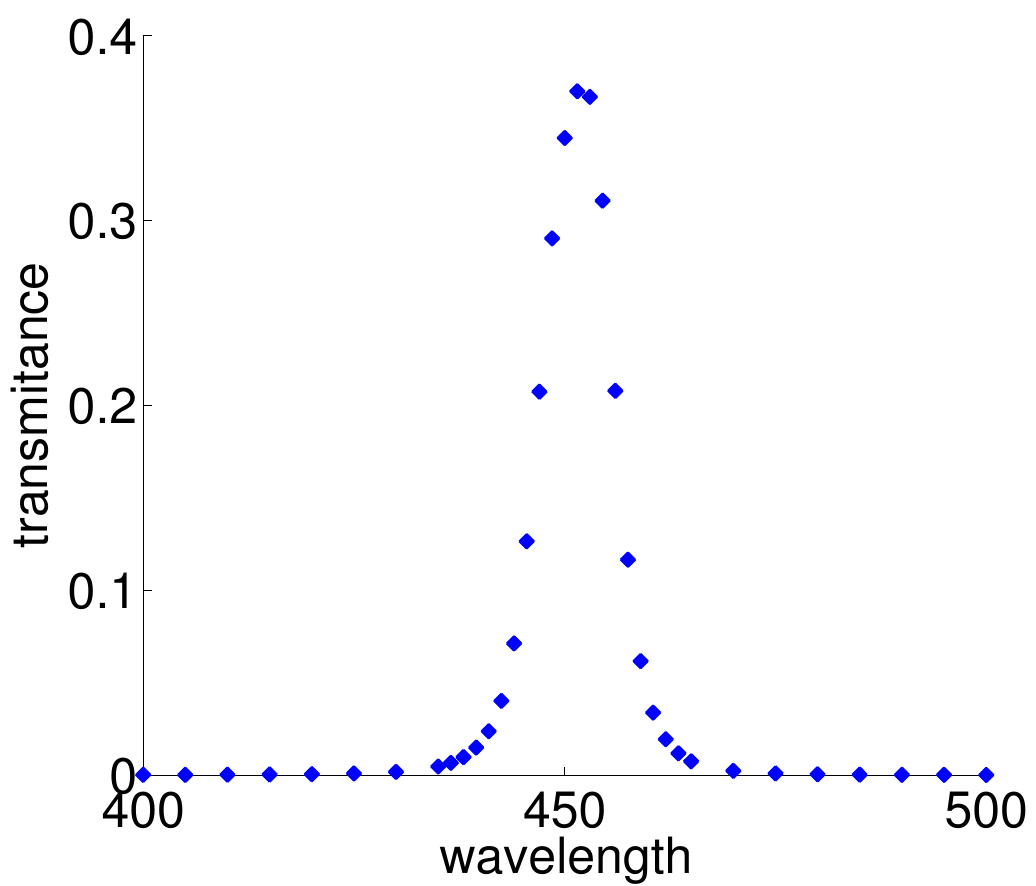}
\includegraphics[width=0.41\linewidth]{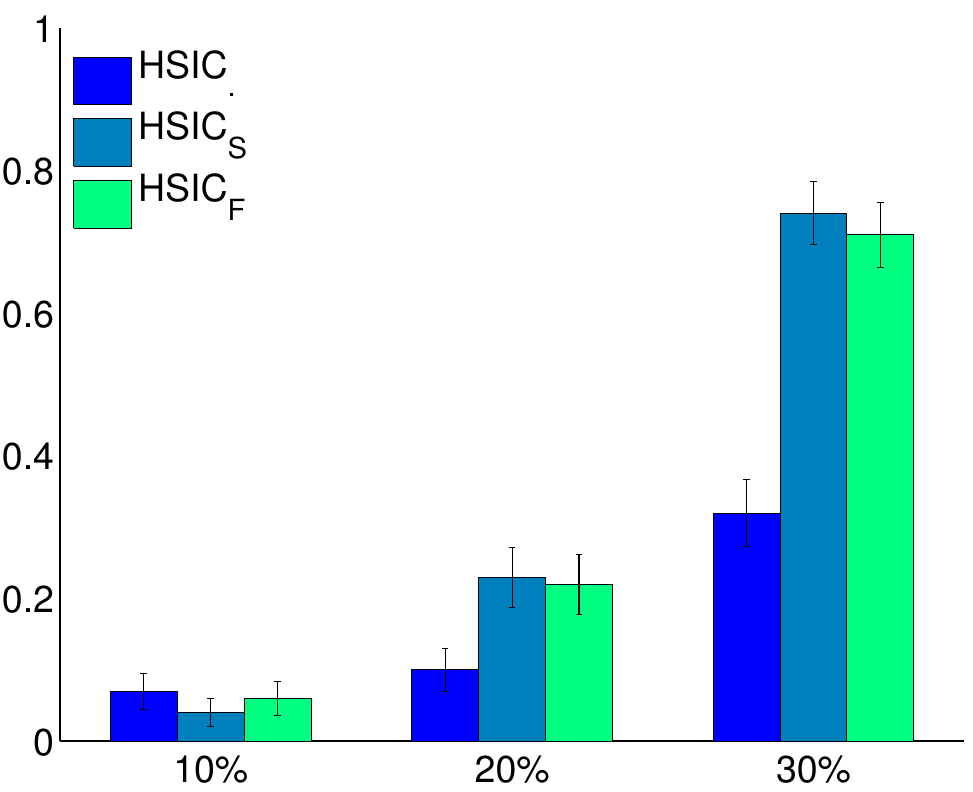}
\includegraphics[width=0.41\linewidth]{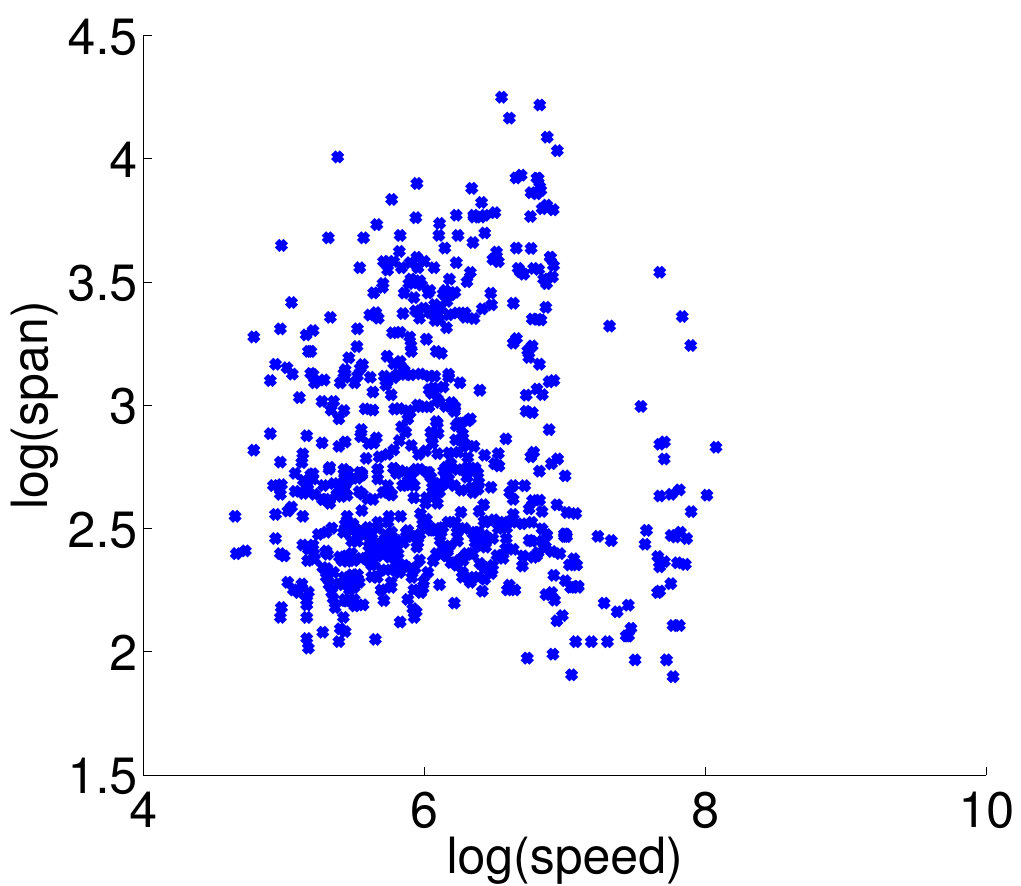}
\includegraphics[width=0.41\linewidth]{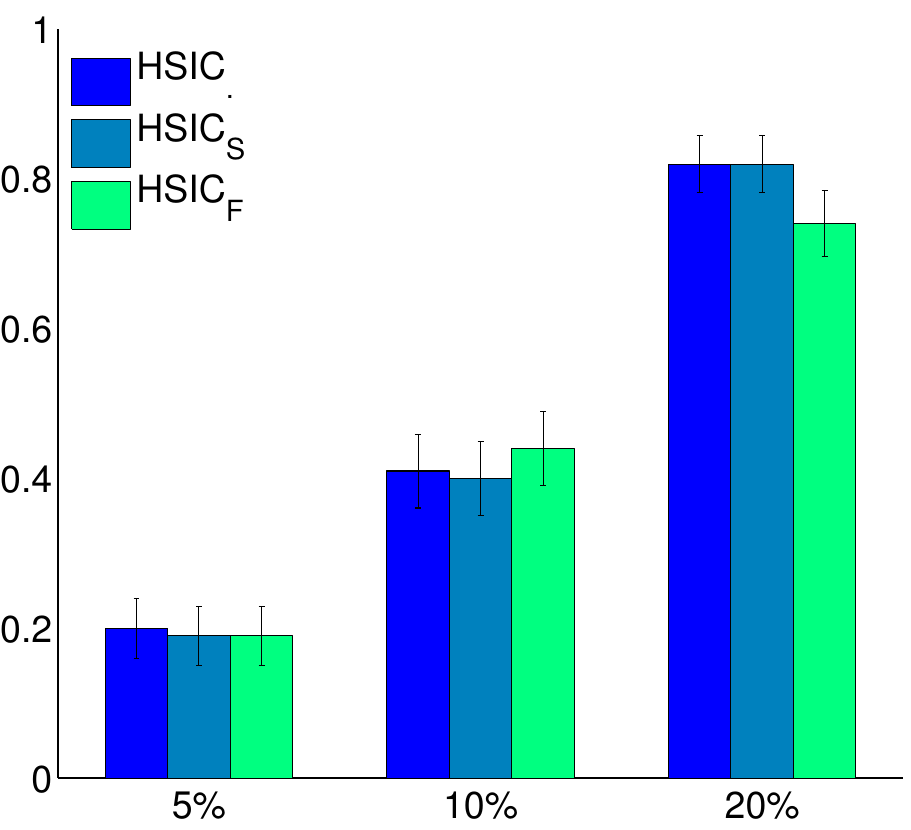}
\caption{Top Row: The left figure shows a plot of wavelength against transmittance. The right figure shows the power of $\HSIC,\HSIC^S,\HSIC^F$ when the data are subsampled to $10\%,20\%,30\%$ (error bars over 100 repetitions). Bottom Row: The left figure shows a plot of $log(wingspan)$ vs $log(airspeed)$. The right figure shows the power of  $\HSIC,\HSIC^S,\HSIC^F$ when the data are subsampled to $5\%,10\%,20\%$ (error bars over 100 repetitions).}
\label{fig:mnist}
\end{figure}

The first is the Eckerle dataset \cite{eckerle} from the NIST Statistical Reference Datasets (NIST StRD) for Nonlinear Regression, data from a NIST study of circular interference transmittance (n=35, $Y$ is transmittance, $X$ is wavelength). A plot of the data in Figure \ref{fig:mnist} reveals a nonlinear relationship between $X,Y$ (though the correlation is 0.035 with p-value 0.84). We subsample the data to see how often we can detect a relationship at $10\%,20\%,30\%$ of the original data size, when the false positive level is always controlled at 0.05. The second is the Aircraft dataset \cite{sm} (n=709, $X$ is log(speed), $Y$ is log(span)). Once again, correlation is low, with a p-value of over 0.8, and we subsample the data to $5\%,10\%,20\%$ of the original data size.

\begin{table}
\hspace{-0.5in}
\begin{tabular}{|c|c|c|c|c|c|c|c|c|c|c|c|c|c|c|c|}
\hline
\multicolumn{1}{|c|}{} &\multicolumn{5}{|c|}{$\alpha$ = 0.01} &\multicolumn{5}{|c|}{$\alpha$ = 0.05} &\multicolumn{5}{|c|}{$\alpha$ = 0.10} \\
\hline
 &\small{$\HSIC$} &\small{$\HSIC_S$} &\small{$\HSIC_F$} & & &\small{$\HSIC$} &\small{$\HSIC_S$} &\small{$\HSIC_F$} & & &\small{$\HSIC$} &\small{$\HSIC_S$} &\small{$\HSIC_F$} & & \\
\hline
\hline\begin{minipage}{.05\textwidth}\includegraphics[width=\linewidth]{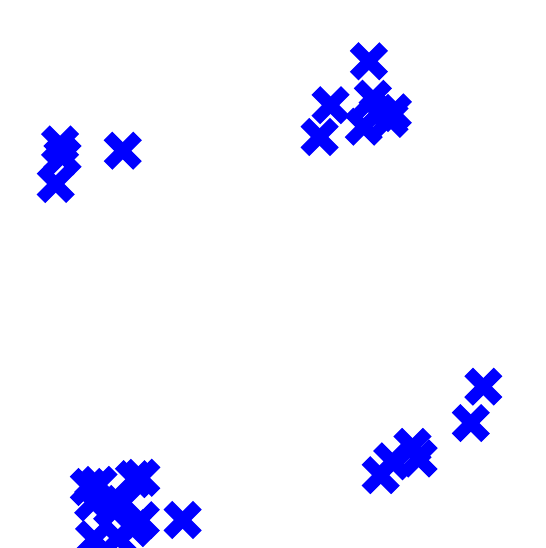}\end{minipage}&\vtop{\hbox{\strut 0.22 }\hbox{\strut {\scriptsize$\pm$0.03}}}&\vtop{\hbox{\strut 0.21 }\hbox{\strut {\scriptsize$\pm$0.03}}}&\vtop{\hbox{\strut 0.34 }\hbox{\strut {\scriptsize$\pm$0.03}}}&&\textcolor{red}{\cmark}&\vtop{\hbox{\strut 0.52 }\hbox{\strut {\scriptsize$\pm$0.04}}}&\vtop{\hbox{\strut 0.52 }\hbox{\strut {\scriptsize$\pm$0.04}}}&\vtop{\hbox{\strut 0.71 }\hbox{\strut {\scriptsize$\pm$0.03}}}&&\textcolor{red}{\cmark}&\vtop{\hbox{\strut 0.73 }\hbox{\strut {\scriptsize$\pm$0.03}}}&\vtop{\hbox{\strut 0.72 }\hbox{\strut {\scriptsize$\pm$0.03}}}&\vtop{\hbox{\strut 0.90 }\hbox{\strut {\scriptsize$\pm$0.02}}}&&\textcolor{red}{\cmark}\\
\hline
\begin{minipage}{.05\textwidth}\includegraphics[width=\linewidth]{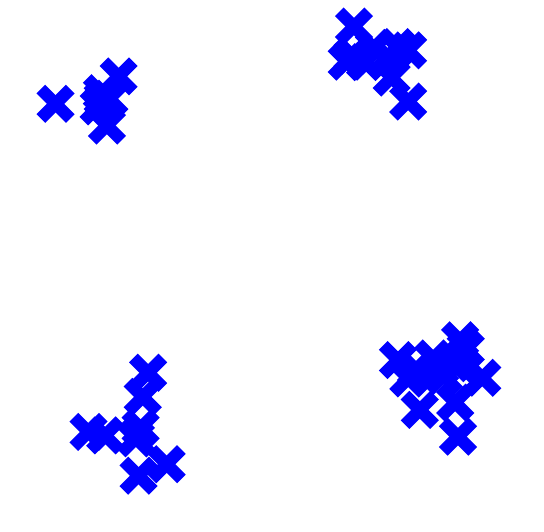}\end{minipage}&\vtop{\hbox{\strut 0.41 }\hbox{\strut {\scriptsize$\pm$0.03}}}&\vtop{\hbox{\strut 0.41 }\hbox{\strut {\scriptsize$\pm$0.03}}}&\vtop{\hbox{\strut 0.48 }\hbox{\strut {\scriptsize$\pm$0.04}}}&&\textcolor{red}{\cmark}&\vtop{\hbox{\strut 0.68 }\hbox{\strut {\scriptsize$\pm$0.03}}}&\vtop{\hbox{\strut 0.68 }\hbox{\strut {\scriptsize$\pm$0.03}}}&\vtop{\hbox{\strut 0.88 }\hbox{\strut {\scriptsize$\pm$0.02}}}&&\textcolor{red}{\cmark}&\vtop{\hbox{\strut 0.85 }\hbox{\strut {\scriptsize$\pm$0.03}}}&\vtop{\hbox{\strut 0.85 }\hbox{\strut {\scriptsize$\pm$0.02}}}&\vtop{\hbox{\strut 0.99 }\hbox{\strut {\scriptsize$\pm$0.01}}}&&\textcolor{red}{\cmark}\\
\hline
\begin{minipage}{.05\textwidth}\includegraphics[width=\linewidth]{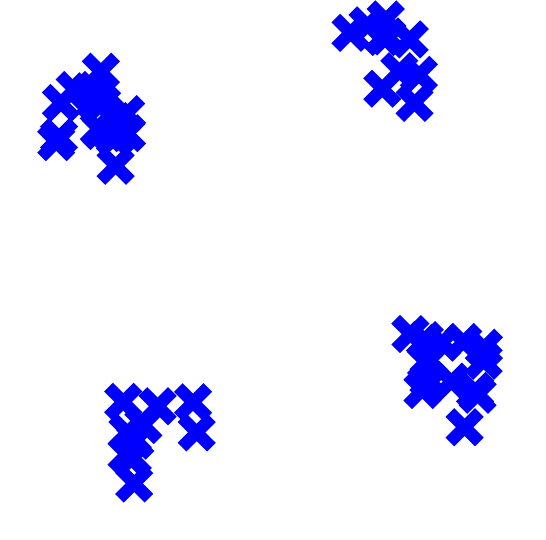}\end{minipage}&\vtop{\hbox{\strut 0.41 }\hbox{\strut {\scriptsize$\pm$0.03}}}&\vtop{\hbox{\strut 0.40 }\hbox{\strut {\scriptsize$\pm$0.03}}}&\vtop{\hbox{\strut 0.52 }\hbox{\strut {\scriptsize$\pm$0.04}}}&&\textcolor{red}{\cmark}&\vtop{\hbox{\strut 0.74 }\hbox{\strut {\scriptsize$\pm$0.03}}}&\vtop{\hbox{\strut 0.74 }\hbox{\strut {\scriptsize$\pm$0.03}}}&\vtop{\hbox{\strut 0.94 }\hbox{\strut {\scriptsize$\pm$0.02}}}&&\textcolor{red}{\cmark}&\vtop{\hbox{\strut 0.94 }\hbox{\strut {\scriptsize$\pm$0.02}}}&\vtop{\hbox{\strut 0.94 }\hbox{\strut {\scriptsize$\pm$0.02}}}&\vtop{\hbox{\strut 0.99 }\hbox{\strut {\scriptsize$\pm$0.01}}}&&\textcolor{red}{\cmark}\\
\hline
\begin{minipage}{.05\textwidth}\includegraphics[width=\linewidth]{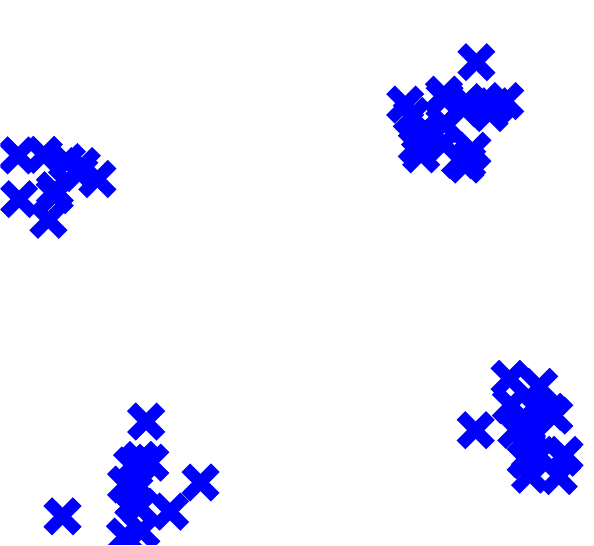}\end{minipage}&\vtop{\hbox{\strut 0.52 }\hbox{\strut {\scriptsize$\pm$0.04}}}&\vtop{\hbox{\strut 0.52 }\hbox{\strut {\scriptsize$\pm$0.04}}}&\vtop{\hbox{\strut 0.66 }\hbox{\strut {\scriptsize$\pm$0.03}}}&&\textcolor{red}{\cmark}&\vtop{\hbox{\strut 0.91 }\hbox{\strut {\scriptsize$\pm$0.02}}}&\vtop{\hbox{\strut 0.91 }\hbox{\strut {\scriptsize$\pm$0.02}}}&\vtop{\hbox{\strut 0.89 }\hbox{\strut {\scriptsize$\pm$0.02}}}&&&\textcolor{blue}{\vtop{\hbox{\strut 0.99 }\hbox{\strut {\scriptsize$\pm$0.01}}}}&\textcolor{blue}{\vtop{\hbox{\strut 0.99 }\hbox{\strut {\scriptsize$\pm$0.01}}}}&\textcolor{blue}{\vtop{\hbox{\strut 0.96 }\hbox{\strut {\scriptsize$\pm$0.01}}}}&&\textcolor{blue}{\xmark}\\
\hline
\hline\begin{minipage}{.05\textwidth}\includegraphics[width=\linewidth]{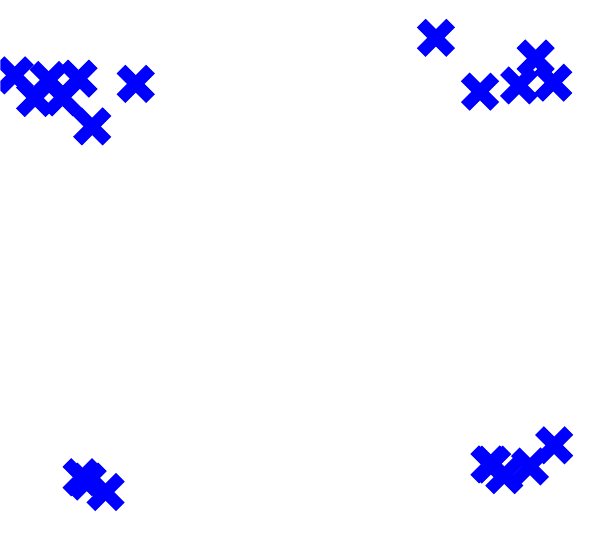}\end{minipage}&\vtop{\hbox{\strut 0.04 }\hbox{\strut {\scriptsize$\pm$0.01}}}&\vtop{\hbox{\strut 0.04 }\hbox{\strut {\scriptsize$\pm$0.01}}}&\vtop{\hbox{\strut 0.04 }\hbox{\strut {\scriptsize$\pm$0.01}}}&&&\vtop{\hbox{\strut 0.12 }\hbox{\strut {\scriptsize$\pm$0.02}}}&\vtop{\hbox{\strut 0.12 }\hbox{\strut {\scriptsize$\pm$0.02}}}&\vtop{\hbox{\strut 0.14 }\hbox{\strut {\scriptsize$\pm$0.02}}}&&&\vtop{\hbox{\strut 0.23 }\hbox{\strut {\scriptsize$\pm$0.03}}}&\vtop{\hbox{\strut 0.23 }\hbox{\strut {\scriptsize$\pm$0.03}}}&\vtop{\hbox{\strut 0.24 }\hbox{\strut {\scriptsize$\pm$0.03}}}&&\\
\hline
\begin{minipage}{.05\textwidth}\includegraphics[width=\linewidth]{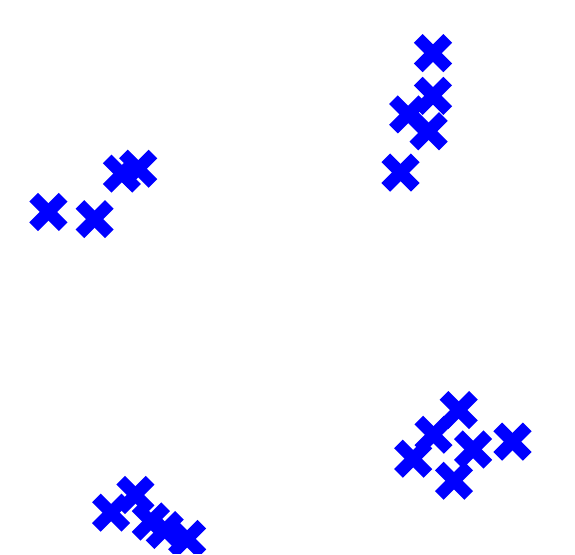}\end{minipage}&\vtop{\hbox{\strut 0.10 }\hbox{\strut {\scriptsize$\pm$0.02}}}&\vtop{\hbox{\strut 0.10 }\hbox{\strut {\scriptsize$\pm$0.02}}}&\vtop{\hbox{\strut 0.12 }\hbox{\strut {\scriptsize$\pm$0.02}}}&&&\vtop{\hbox{\strut 0.31 }\hbox{\strut {\scriptsize$\pm$0.03}}}&\vtop{\hbox{\strut 0.31 }\hbox{\strut {\scriptsize$\pm$0.03}}}&\vtop{\hbox{\strut 0.40 }\hbox{\strut {\scriptsize$\pm$0.03}}}&&\textcolor{red}{\cmark}&\vtop{\hbox{\strut 0.47 }\hbox{\strut {\scriptsize$\pm$0.04}}}&\vtop{\hbox{\strut 0.47 }\hbox{\strut {\scriptsize$\pm$0.04}}}&\vtop{\hbox{\strut 0.58 }\hbox{\strut {\scriptsize$\pm$0.03}}}&&\textcolor{red}{\cmark}\\
\hline
\begin{minipage}{.05\textwidth}\includegraphics[width=\linewidth]{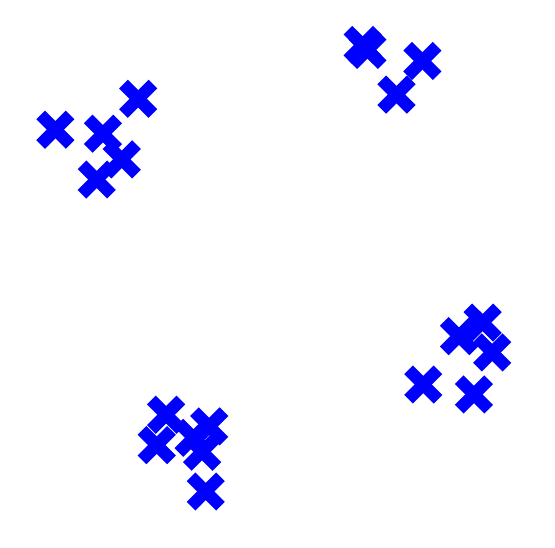}\end{minipage}&\vtop{\hbox{\strut 0.33 }\hbox{\strut {\scriptsize$\pm$0.03}}}&\vtop{\hbox{\strut 0.33 }\hbox{\strut {\scriptsize$\pm$0.03}}}&\vtop{\hbox{\strut 0.46 }\hbox{\strut {\scriptsize$\pm$0.04}}}&&\textcolor{red}{\cmark}&\vtop{\hbox{\strut 0.77 }\hbox{\strut {\scriptsize$\pm$0.03}}}&\vtop{\hbox{\strut 0.77 }\hbox{\strut {\scriptsize$\pm$0.03}}}&\vtop{\hbox{\strut 0.91 }\hbox{\strut {\scriptsize$\pm$0.02}}}&&\textcolor{red}{\cmark}&\vtop{\hbox{\strut 0.95 }\hbox{\strut {\scriptsize$\pm$0.01}}}&\vtop{\hbox{\strut 0.96 }\hbox{\strut {\scriptsize$\pm$0.01}}}&\vtop{\hbox{\strut 0.99 }\hbox{\strut {\scriptsize$\pm$0.01}}}&&\textcolor{red}{\cmark}\\
\hline
\begin{minipage}{.05\textwidth}\includegraphics[width=\linewidth]{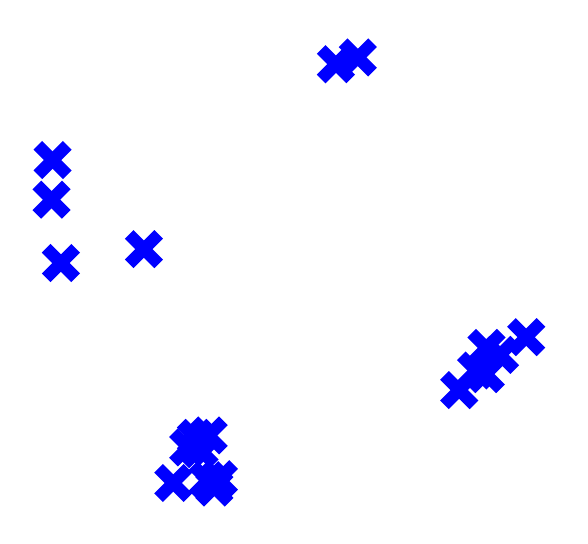}\end{minipage}&\vtop{\hbox{\strut 0.93 }\hbox{\strut {\scriptsize$\pm$0.02}}}&\vtop{\hbox{\strut 0.93 }\hbox{\strut {\scriptsize$\pm$0.02}}}&\vtop{\hbox{\strut 0.96 }\hbox{\strut {\scriptsize$\pm$0.01}}}&&\textcolor{red}{\cmark}&\textcolor{red}{\vtop{\hbox{\strut 1.00 }\hbox{\strut {\scriptsize$\pm$0.00}}}}&\textcolor{red}{\vtop{\hbox{\strut 1.00 }\hbox{\strut {\scriptsize$\pm$0.00}}}}&\textcolor{red}{\vtop{\hbox{\strut 1.00 }\hbox{\strut {\scriptsize$\pm$0.00}}}}&&&\textcolor{red}{\vtop{\hbox{\strut 1.00 }\hbox{\strut {\scriptsize$\pm$0.00}}}}&\textcolor{red}{\vtop{\hbox{\strut 1.00 }\hbox{\strut {\scriptsize$\pm$0.00}}}}&\textcolor{red}{\vtop{\hbox{\strut 1.00 }\hbox{\strut {\scriptsize$\pm$0.00}}}}&&\\
\hline
\hline\begin{minipage}{.05\textwidth}\includegraphics[width=\linewidth]{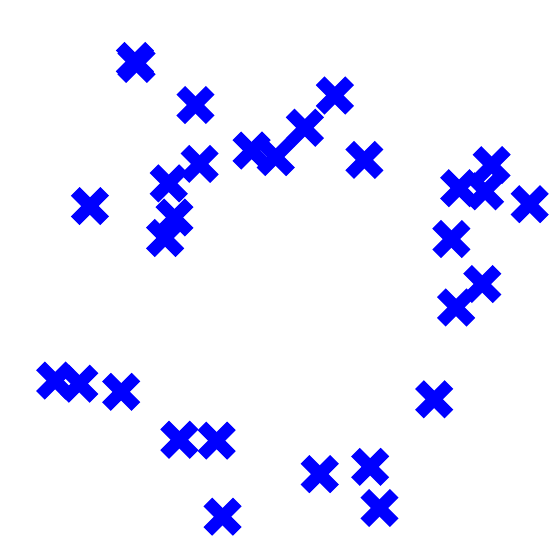}\end{minipage}&\vtop{\hbox{\strut 0.07 }\hbox{\strut {\scriptsize$\pm$0.02}}}&\vtop{\hbox{\strut 0.07 }\hbox{\strut {\scriptsize$\pm$0.02}}}&\vtop{\hbox{\strut 0.09 }\hbox{\strut {\scriptsize$\pm$0.02}}}&&&\vtop{\hbox{\strut 0.24 }\hbox{\strut {\scriptsize$\pm$0.03}}}&\vtop{\hbox{\strut 0.26 }\hbox{\strut {\scriptsize$\pm$0.03}}}&\vtop{\hbox{\strut 0.32 }\hbox{\strut {\scriptsize$\pm$0.03}}}&&\textcolor{red}{\cmark}&\vtop{\hbox{\strut 0.44 }\hbox{\strut {\scriptsize$\pm$0.04}}}&\vtop{\hbox{\strut 0.47 }\hbox{\strut {\scriptsize$\pm$0.04}}}&\vtop{\hbox{\strut 0.48 }\hbox{\strut {\scriptsize$\pm$0.04}}}&&\\
\hline
\begin{minipage}{.05\textwidth}\includegraphics[width=\linewidth]{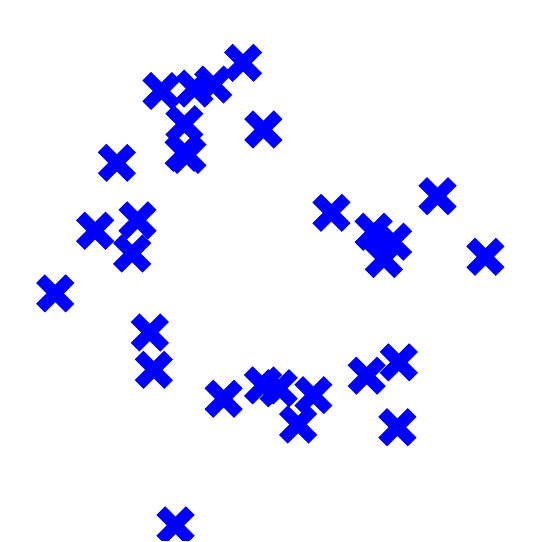}\end{minipage}&\vtop{\hbox{\strut 0.06 }\hbox{\strut {\scriptsize$\pm$0.02}}}&\vtop{\hbox{\strut 0.07 }\hbox{\strut {\scriptsize$\pm$0.02}}}&\vtop{\hbox{\strut 0.09 }\hbox{\strut {\scriptsize$\pm$0.02}}}&&&\vtop{\hbox{\strut 0.26 }\hbox{\strut {\scriptsize$\pm$0.03}}}&\vtop{\hbox{\strut 0.28 }\hbox{\strut {\scriptsize$\pm$0.03}}}&\vtop{\hbox{\strut 0.32 }\hbox{\strut {\scriptsize$\pm$0.03}}}&&&\vtop{\hbox{\strut 0.45 }\hbox{\strut {\scriptsize$\pm$0.04}}}&\vtop{\hbox{\strut 0.47 }\hbox{\strut {\scriptsize$\pm$0.04}}}&\vtop{\hbox{\strut 0.48 }\hbox{\strut {\scriptsize$\pm$0.04}}}&&\\
\hline
\begin{minipage}{.05\textwidth}\includegraphics[width=\linewidth]{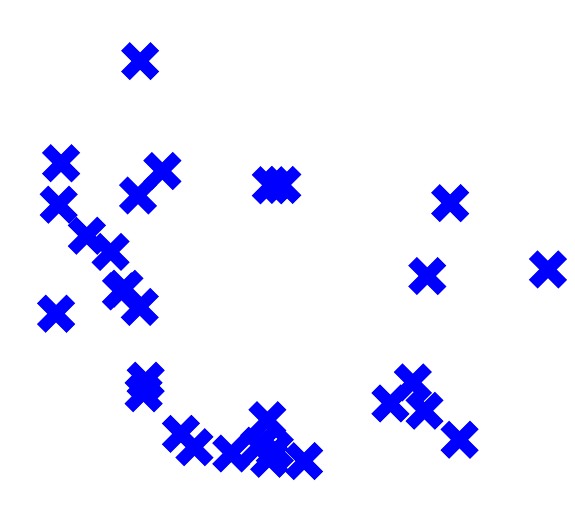}\end{minipage}&\vtop{\hbox{\strut 0.10 }\hbox{\strut {\scriptsize$\pm$0.02}}}&\vtop{\hbox{\strut 0.12 }\hbox{\strut {\scriptsize$\pm$0.02}}}&\vtop{\hbox{\strut 0.14 }\hbox{\strut {\scriptsize$\pm$0.02}}}&&&\vtop{\hbox{\strut 0.34 }\hbox{\strut {\scriptsize$\pm$0.03}}}&\vtop{\hbox{\strut 0.34 }\hbox{\strut {\scriptsize$\pm$0.03}}}&\vtop{\hbox{\strut 0.39 }\hbox{\strut {\scriptsize$\pm$0.03}}}&&&\vtop{\hbox{\strut 0.51 }\hbox{\strut {\scriptsize$\pm$0.04}}}&\vtop{\hbox{\strut 0.52 }\hbox{\strut {\scriptsize$\pm$0.04}}}&\vtop{\hbox{\strut 0.53 }\hbox{\strut {\scriptsize$\pm$0.04}}}&&\\
\hline
\begin{minipage}{.05\textwidth}\includegraphics[width=\linewidth]{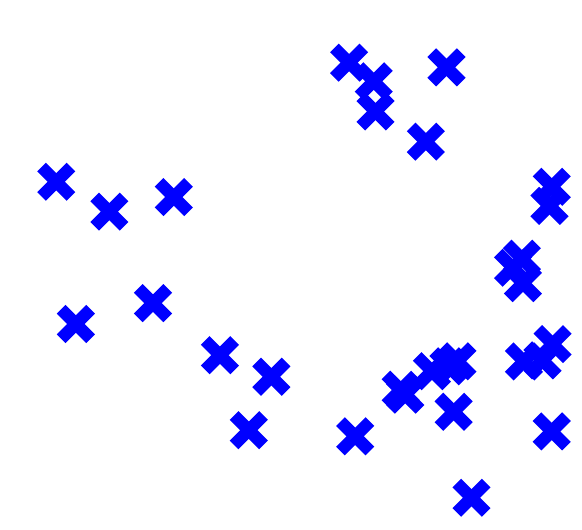}\end{minipage}&\vtop{\hbox{\strut 0.07 }\hbox{\strut {\scriptsize$\pm$0.02}}}&\vtop{\hbox{\strut 0.07 }\hbox{\strut {\scriptsize$\pm$0.02}}}&\vtop{\hbox{\strut 0.10 }\hbox{\strut {\scriptsize$\pm$0.02}}}&&\textcolor{red}{\cmark}&\vtop{\hbox{\strut 0.30 }\hbox{\strut {\scriptsize$\pm$0.03}}}&\vtop{\hbox{\strut 0.33 }\hbox{\strut {\scriptsize$\pm$0.03}}}&\vtop{\hbox{\strut 0.35 }\hbox{\strut {\scriptsize$\pm$0.03}}}&&&\vtop{\hbox{\strut 0.53 }\hbox{\strut {\scriptsize$\pm$0.04}}}&\vtop{\hbox{\strut 0.54 }\hbox{\strut {\scriptsize$\pm$0.04}}}&\vtop{\hbox{\strut 0.57 }\hbox{\strut {\scriptsize$\pm$0.04}}}&&\\
\hline
\hline\begin{minipage}{.05\textwidth}\includegraphics[width=\linewidth]{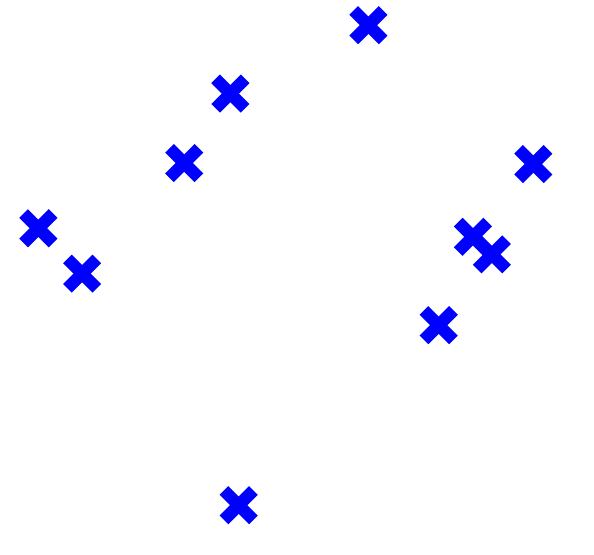}\end{minipage}&\vtop{\hbox{\strut 0.04 }\hbox{\strut {\scriptsize$\pm$0.01}}}&\vtop{\hbox{\strut 0.05 }\hbox{\strut {\scriptsize$\pm$0.02}}}&\vtop{\hbox{\strut 0.04 }\hbox{\strut {\scriptsize$\pm$0.01}}}&&&\vtop{\hbox{\strut 0.18 }\hbox{\strut {\scriptsize$\pm$0.03}}}&\vtop{\hbox{\strut 0.27 }\hbox{\strut {\scriptsize$\pm$0.03}}}&\vtop{\hbox{\strut 0.24 }\hbox{\strut {\scriptsize$\pm$0.03}}}&\cmark&\textcolor{red}{\cmark}&\vtop{\hbox{\strut 0.34 }\hbox{\strut {\scriptsize$\pm$0.03}}}&\vtop{\hbox{\strut 0.45 }\hbox{\strut {\scriptsize$\pm$0.04}}}&\vtop{\hbox{\strut 0.44 }\hbox{\strut {\scriptsize$\pm$0.04}}}&\cmark&\textcolor{red}{\cmark}\\
\hline
\begin{minipage}{.05\textwidth}\includegraphics[width=\linewidth]{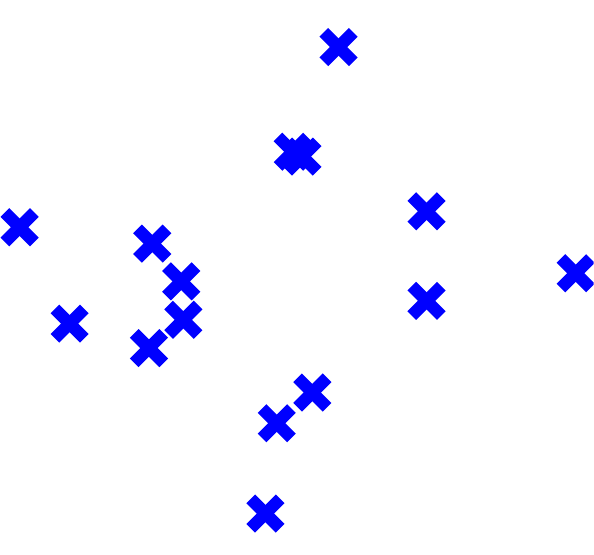}\end{minipage}&\vtop{\hbox{\strut 0.16 }\hbox{\strut {\scriptsize$\pm$0.03}}}&\vtop{\hbox{\strut 0.20 }\hbox{\strut {\scriptsize$\pm$0.03}}}&\vtop{\hbox{\strut 0.20 }\hbox{\strut {\scriptsize$\pm$0.03}}}&&&\vtop{\hbox{\strut 0.45 }\hbox{\strut {\scriptsize$\pm$0.04}}}&\vtop{\hbox{\strut 0.58 }\hbox{\strut {\scriptsize$\pm$0.03}}}&\vtop{\hbox{\strut 0.58 }\hbox{\strut {\scriptsize$\pm$0.03}}}&\cmark&\textcolor{red}{\cmark}&\vtop{\hbox{\strut 0.67 }\hbox{\strut {\scriptsize$\pm$0.03}}}&\vtop{\hbox{\strut 0.73 }\hbox{\strut {\scriptsize$\pm$0.03}}}&\vtop{\hbox{\strut 0.73 }\hbox{\strut {\scriptsize$\pm$0.03}}}&\cmark&\textcolor{red}{\cmark}\\
\hline
\begin{minipage}{.05\textwidth}\includegraphics[width=\linewidth]{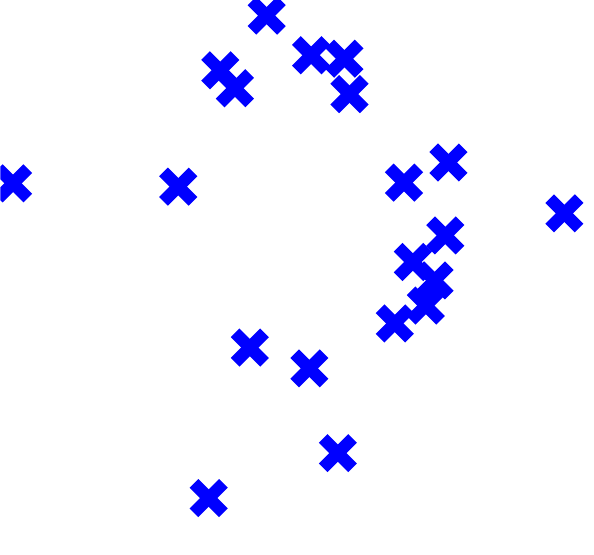}\end{minipage}&\vtop{\hbox{\strut 0.34 }\hbox{\strut {\scriptsize$\pm$0.03}}}&\vtop{\hbox{\strut 0.43 }\hbox{\strut {\scriptsize$\pm$0.04}}}&\vtop{\hbox{\strut 0.43 }\hbox{\strut {\scriptsize$\pm$0.04}}}&\cmark&\textcolor{red}{\cmark}&\vtop{\hbox{\strut 0.71 }\hbox{\strut {\scriptsize$\pm$0.03}}}&\vtop{\hbox{\strut 0.80 }\hbox{\strut {\scriptsize$\pm$0.03}}}&\vtop{\hbox{\strut 0.79 }\hbox{\strut {\scriptsize$\pm$0.03}}}&\cmark&\textcolor{red}{\cmark}&\vtop{\hbox{\strut 0.85 }\hbox{\strut {\scriptsize$\pm$0.03}}}&\vtop{\hbox{\strut 0.90 }\hbox{\strut {\scriptsize$\pm$0.02}}}&\vtop{\hbox{\strut 0.89 }\hbox{\strut {\scriptsize$\pm$0.02}}}&\cmark&\\
\hline
\begin{minipage}{.05\textwidth}\includegraphics[width=\linewidth]{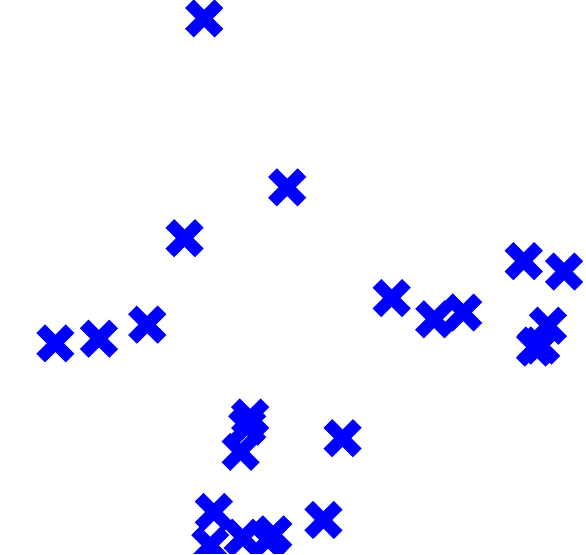}\end{minipage}&\vtop{\hbox{\strut 0.63 }\hbox{\strut {\scriptsize$\pm$0.03}}}&\vtop{\hbox{\strut 0.72 }\hbox{\strut {\scriptsize$\pm$0.03}}}&\vtop{\hbox{\strut 0.73 }\hbox{\strut {\scriptsize$\pm$0.03}}}&\cmark&\textcolor{red}{\cmark}&\vtop{\hbox{\strut 0.91 }\hbox{\strut {\scriptsize$\pm$0.02}}}&\vtop{\hbox{\strut 0.92 }\hbox{\strut {\scriptsize$\pm$0.02}}}&\vtop{\hbox{\strut 0.92 }\hbox{\strut {\scriptsize$\pm$0.02}}}&&&\vtop{\hbox{\strut 0.95 }\hbox{\strut {\scriptsize$\pm$0.01}}}&\vtop{\hbox{\strut 0.96 }\hbox{\strut {\scriptsize$\pm$0.01}}}&\vtop{\hbox{\strut 0.96 }\hbox{\strut {\scriptsize$\pm$0.01}}}&&\\
\hline
\end{tabular}
\caption{
The first column shows scatterplots of $X$ vs $Y$ (all having dependence between $X,Y$). There are 3 sets of 5 columns each - for $\alpha=0.01, 0.05, 0.1$ (controlled by running 2000 permutations). In eachs set, the first three columns show the power of $\HSIC,\HSIC^S,\HSIC^F$ (with standard deviation over 200 repetitions below). The fourth column shows when $\HSIC^S$ is significantly better than $\HSIC$, and the fifth column when $\HSIC^F$ has significantly higher power than $\HSIC$. A blank means the powers are not significantly better or worse. In the first dataset (A) (top 4) we show how the power varies with increasing $n$ (becomes easier). In the second dataset (B) (second 4) we show how the power varies with rotation (goes from near-independence to clear dependence). In the third dataset (C) (third 4), we demonstrate a case where shrinkage does \textit{not} help much, which is a circle with a hole. In the last dataset (D) (last 4), we demonstrate a case where $\HSIC^S$ does as well as $\HSIC^F$. We tried many more datasets, these are a few representative samples.
}
\label{tab:synth}
\end{table}
\clearpage


\section{Discussion}

\begin{figure} [h]
\centering
\includegraphics[width=0.6\linewidth]{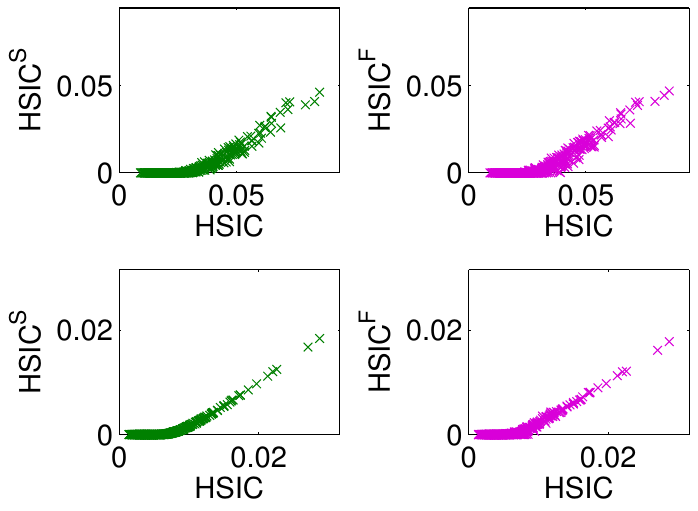}
\caption{The top row corresponds to $n=20$, and the bottom row has $n=50$. The left plots compare $\HSIC^S$ to $\HSIC$, and the right plots compare $\HSIC^F$ to $\HSIC$. Each cross mark corresponds to the shrunk and unshrunk HSIC calculated during a single permutation of a permutation test.}
\label{fig:hsicscatter}
\end{figure}

Why might shrinkage improve power? Let us examine the net effect of using shrunk estimators on the value of HSIC, i.e. let us compare $\HSIC^S$ and $\HSIC^F$ to  $\HSIC$ by computing these over all the repetitions of the permutation testing procedure described in the introduction.
In Fig. \ref{fig:hsicscatter}, both estimators are visually similar in transforming the actual test statistic. Perhaps the more interesting phenomenon is that Fig. \ref{fig:hsicscatter} is reminiscent of the graph of a soft-thresholding operator $ST_t(x) = \max\{0,x-t\}$. Intuitively, if the unshrunk HSIC value is small, the shrinkage methods deem it to be ``noise'' and it is shrunk to zero. Looking at the X-axis scaling of the top and bottom row, the size of the region that gets shrunk to zero decreases with $n$ - as expected, shrinkage has less effect when $\SXY$ has low variance). The shrinkage being non-monotone (more so for $n=20$ than $n=50$ in Figure \ref{fig:hsicscatter}) is key to achieving an improvement in power. 

Using the  intuition from the above figure, we can finally piece together why shrinkage may  yield benefits. A rejection of $\Hc_0$ occurs when the test statistic stands out in the right tail of its null distribution. Typically, when the alternative is true (this is when rejecting the null improves power) the unshrunk test statistics calculated from the permuted samples is smaller than the unshrunk HSIC calculated on the original sample. However, the effect of shrinking the small statistics towards zero, and setting the smallest ones to zero, is that the unpermuted test statistic under the alternative distribution stands out more in the right tail of the null.

\begin{figure} [h]
\centering
\includegraphics[width=0.41\linewidth]{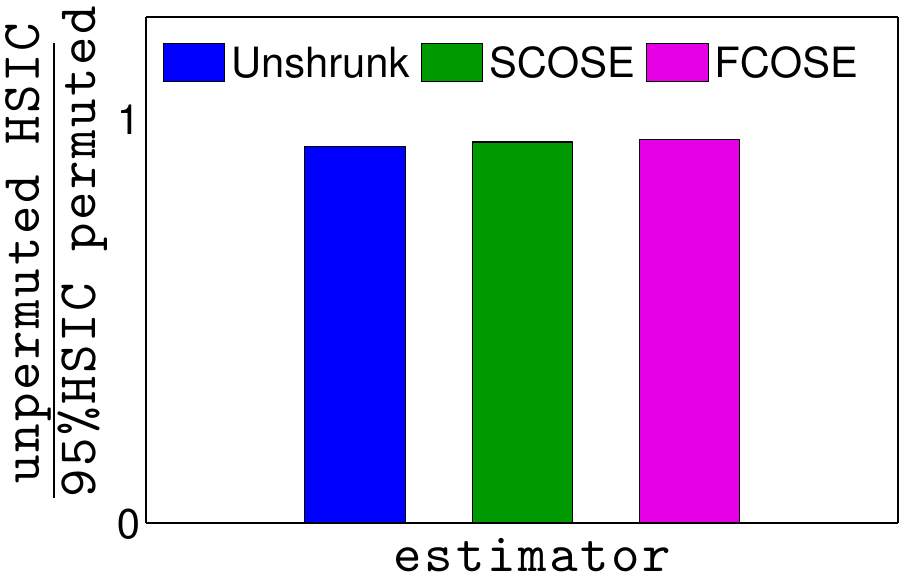}
\includegraphics[width=0.41\linewidth]{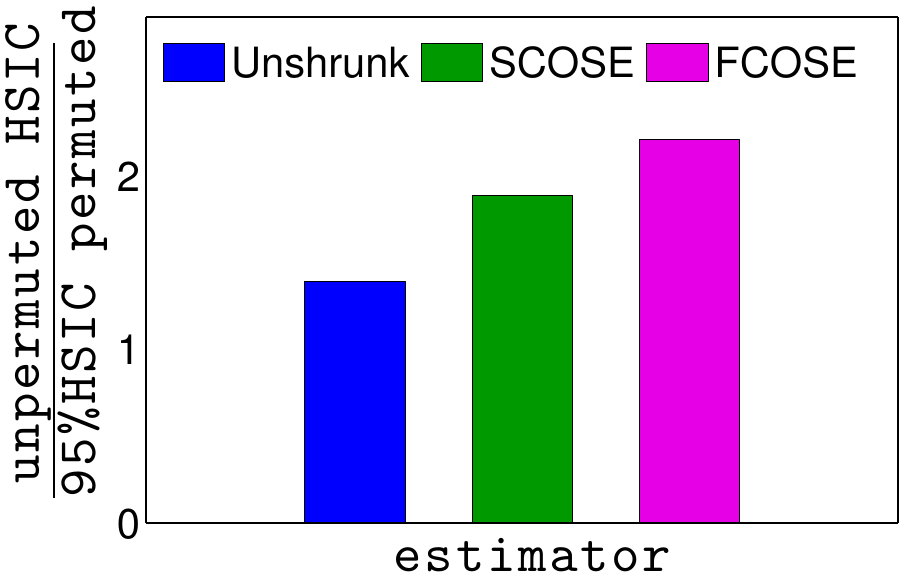}
\includegraphics[width=0.41\linewidth]{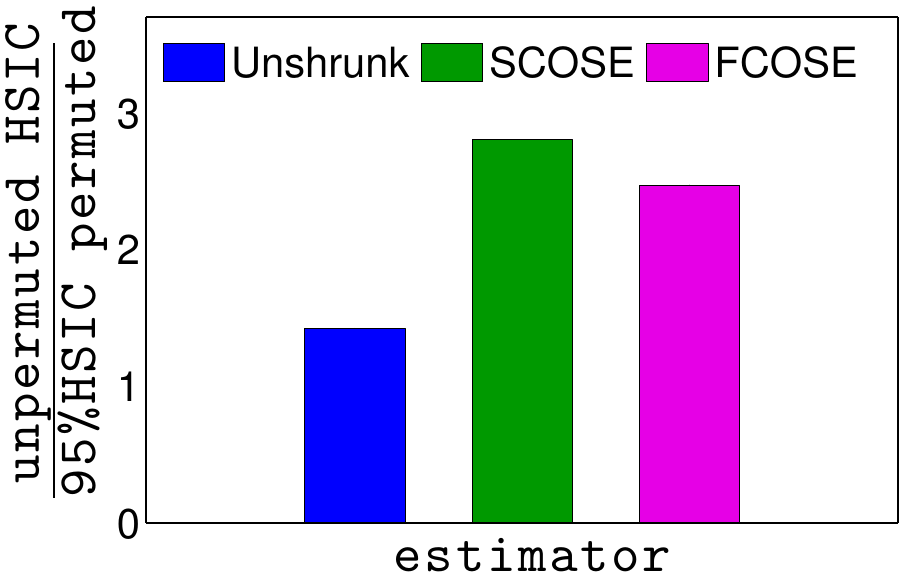}
\includegraphics[width=0.41\linewidth]{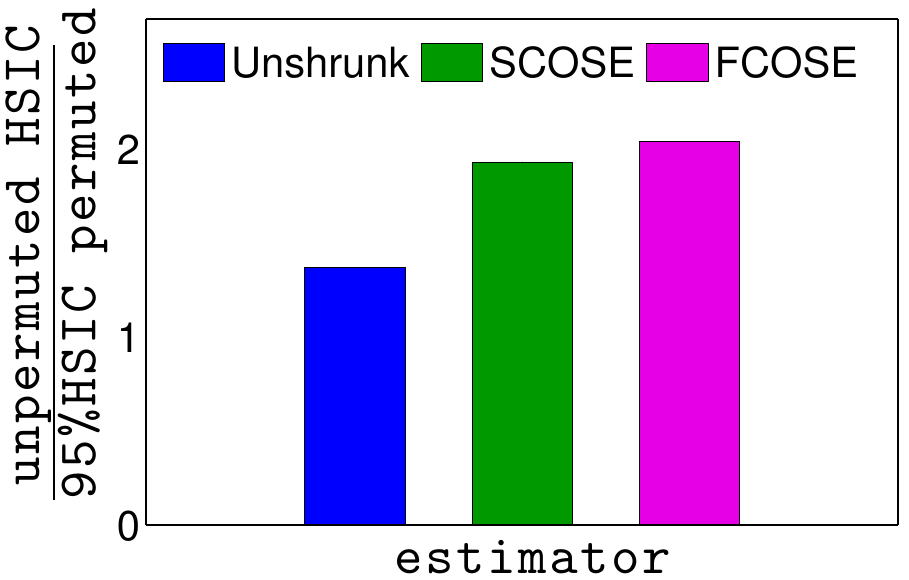}
\caption{All panels show the ratio of the unpermuted HSIC to the 95{th} percentile of the null distribution based on HSICs calculated from the permuted data. (see Table \ref{tab:synth}) The top row has datasets (C) with radius 2.2, (B) with angle $3 \times \pi/32$, and the bottom row has (D) with $N=25$, (A) with $N=40$. These observations were qualitatively the same in all other synthetic data parameter settings, and also for other percentiles than $95$th, and since the figures look  identical in spirit, they were omitted due to lack of space.}
\label{fig:ratio}
\end{figure}

In other words, relative to the unshrunk null distribution and the unshrunk test statistic, the tail of the null distribution is shrunk more towards zero than the unpermuted test statistic, causing the latter to have a higher quantile in the right tail of the former (relative to the quantile before shrinkage). 
Let us verify this experimentally. In  Fig.\ref{fig:ratio} we plot for each of the datasets in Table \ref{tab:synth}, the average ratio of unpermuted statistic T to the 95th percentile of the permuted statistics, for $T \in \{\HSIC, \HSIC^S, \HSIC^F\}$. Recall that for dataset (C), we didn't see much of an improvement in power, but for (A),(B),(D) it is clear from Fig. \ref{fig:ratio} that the unpermuted statistic is shrunk less than its null distribution's 95th quantile.

\textbf{Remark.} In our experiments, real and synthetic, shrinkage usually improves (and almost never worsens) power in false-positive regimes that we usually care about. Will shrinkage \textit{always} improve power? Possibly not. 
Even though shrunk the shrunk $\SXY$ dominates $\SXY$ for estimation error, it may not be the case that shrunk $\HSIC$ always dominates unshrunk $\HSIC$ for test power (i.e. the latter may not be \textit{inadmissible}). However, just as no single classifier always outperforms another, it is still beneficial to add techniques like shrinkage, that seem to consistently yield benefits in practice, to the practitioner's array of tools.

\section{Conclusion}

We presented evidence for an important phenomenon - using biased but lower variance shrunk estimators of cross-covariance operators can often significantly improve test power of HSIC at small sample sizes. This observation (that shrinkage can improve power) has rarely been made in the statistics and machine learning testing literature. We think the reason is that most test statistics for independence testing cannot be immediately expressed as the norm of an empirical operator, making it less obvious \textit{how} to apply shrinkage to improve their power at low sample sizes.

 We also showed the optimality (among linear shrinkage estimators) of SCOSE, but observe that the nonlinear shrinkage of FCOSE usually yields higher power. To the best of our knowledge, there seems to be no current literature showing that the choice made by leave-one-out cross-validation (SCOSE) explicitly leads to an estimator that is "optimal" in some sense (among linear shrinkage estimators). This may be because it is often not possible to explicitly calculate the form of the LOOCV estimator, nor the explicit form of the best linear shrinkage estimator, as can both be done in this simple setting.

 Since even the best possible linear shrinkage estimator (as represented by SCOSE) is usually worse than FCOSE, this result indicates that in order to  improve upon FCOSE, it will be  necessary to further study the class of non-linear shrinkage estimators for our infinite dimensional operators, as done for finite dimensional covariance matrices in \cite{ledoitwolfnonlinear} and other papers by the same authors. 
 
 We ended with a brief investigation into the effect of shrinkage on HSIC and why shrinkage may intuitively improve power. We think that our work will be important for more powerful nonparametric detection of subtle nonlinear dependencies at low sample sizes, a common problem in scientific applications.

\subsection*{Acknowledgments}

We would like to thank Arthur Gretton and Jessica Chemali for useful feedback on an earlier draft of the paper, Larry Wasserman for pointing out some useful references, and Krikamol Muandet for sharing his code. AR was supported in part by ONR MURI grant N000140911052. 

\bibliographystyle{agsm}
\bibliography{story}

\end{document}